\documentclass[11pt]{article}
\usepackage[final]{acl}
\usepackage{times}
\usepackage{latexsym}
\usepackage{amsmath}
\usepackage{amssymb}
\usepackage{booktabs}
\usepackage{graphicx}
\usepackage{multirow}
\usepackage{xcolor}
\usepackage{subcaption}
\usepackage{enumitem}
\usepackage{float}

\title{How Many Pixels Is a Digit Worth? Place-Aware Coordinate Entropy for GUI Agent Confidence Estimation}

\author{
  Yunxiang Li$^{\heartsuit}$,
  Xixin Wu$^{\heartsuit}$,
  Helen Meng$^{\heartsuit}$\thanks{$\;\;$Corresponding author.} \\
  $^\heartsuit$The Chinese University of Hong Kong, Hong Kong SAR, China \\
      \texttt{yli@se.cuhk.edu.hk}
  }

\begin{document}
\maketitle

\begin{abstract}
  GUI agents predict click coordinates as digit-token sequences, but standard text-LLM confidence estimation methods rank correct clicks from wrong ones only weakly.
  GUI-specific alternatives use $K$ samples or new supervision, but still leave room for improvement. We trace part of this to \emph{place-value asymmetry}: bounding-
  box correctness often makes higher-place digits more important than lower-place digits, so uniform aggregation weakens the signal that determines correctness. The
  fix is to weight each digit's Shannon entropy by its place value. We call this \textbf{Place-Aware Coordinate Entropy} (PACE). Across fixed-scale agents on
  ScreenSpot-Pro and ScreenSpot-v$2$, PACE wins both AUROC and selective accuracy on all primary comparisons in a single forward pass, matching or outperforming
  $K$-sample baselines at a fraction of the cost. PACE provides a per-click confidence estimate that turns coordinate-token internals into a practical confidence
signal for GUI agent deployment.
\end{abstract}
\section{Introduction}
\label{sec:intro}

\begin{figure*}[!t]
\centering
\includegraphics[width=0.88\textwidth]{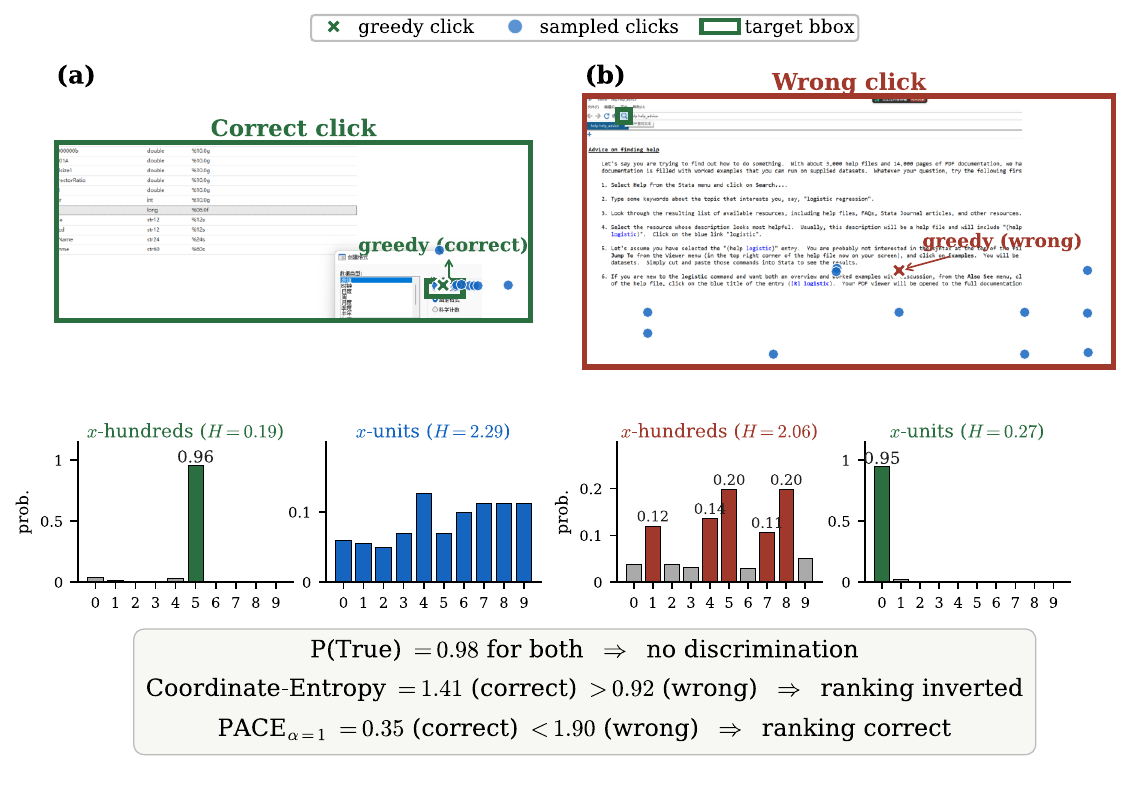}
\caption{\textbf{Place-value asymmetry on two real EvoCUA-$8$B predictions (Stata, ScreenSpot-Pro).} (a) A correct click whose entropy lives in the units digit ($H{=}2.29$, near-uniform over ten candidates $1$ pixel apart) while the hundreds digit is locked ($H{=}0.19$). (b) A wrong click whose entropy lives in the hundreds digit ($H{=}2.06$, five candidates $100$ pixels apart) while the units digit is locked ($H{=}0.27$). Both cases share P(True) ${=}0.98$. Coordinate-Entropy averages uniformly and ranks the
correct click as \emph{more} uncertain ($1.41 {>} 0.92$). PACE weights by place value and recovers the correct ranking ($0.35 {<} 1.90$).}
\label{fig:teaser}
\end{figure*}

Open-source GUI agents predict pixel coordinates of clicks,
keypresses, and scrolls from screenshots, driving professional
desktop tasks end-to-end~\citep{uitars2025,evocua2026,cogagent2024,aguvis2024,osatlas2024}.
Click accuracy on dense professional interfaces remains below half
for the strongest open agents ($35$--$50\%$ on ScreenSpot-Pro~\citep{screenspotpro2025}), and
a wrong click is often irreversible: a sent payment or a deleted
file cannot be undone by resampling~\citep{trustworthygui2025}.
To manage this risk, the grounding model must therefore expose a
  per-prediction confidence estimate that ranks correct clicks above
  incorrect ones, and ideally produces a calibrated probability for
  downstream decisions. The same estimate can guide when the agent should
  act, ask for help, or use extra computation, and can be combined across
  steps in longer tasks.

For ordinary text generation, confidence estimation is a mature
toolkit. Verbalized
prompts~\citep{tian2023calibration,kadavath2022language},
self-consistency over multiple
samples~\citep{wang2023selfconsistency}, and
log-probability~\citep{malinin2021uncertainty} all give usable
confidence signals on factual question answering, semantic parsing,
and reasoning traces. GUI click prediction also generates its output token by token: the
  model outputs a short digit-token sequence that decodes to screen
  coordinates. We find that these signals often provide weak or
  inconsistent discrimination for coordinate outputs
  (Table~\ref{tab:headline}).

We trace part of this to a common mechanism. GUI click correctness is bounding-box-defined: any pixel inside the target counts as correct. The hundreds digit of a coordinate therefore corresponds to larger, often UI-element-level shifts (a $1$-unit confusion is a $100$-pixel error that can land on a neighbor), while the units digit moves the click by only $1$ pixel and is more likely to remain inside the same target. Per-token uncertainty is therefore asymmetric in its effect: entropy at higher-place digits is generally more informative for correctness, although lower-place digits still matter in some cases.
  Common token-level confidence signals aggregate uniformly across this
  asymmetric sequence, weakening the signal related to correctness. We
  call this property \emph{place-value asymmetry}: per-position information
  content of a coordinate token sequence is non-uniform.

We therefore scale each per-digit Shannon entropy by its place value before averaging.  We call this \textbf{Place-Aware Coordinate
Entropy} (PACE), parameterized by a single exponent $\alpha$ on the
place value. For a 3-digit coordinate at $\alpha{=}1$, the hundred
digit's entropy contributes $100\times$ more than the unit digit's to the
score, matching the expected pixel cost of a one-digit error at each
place; $\alpha{=}0$ reduces to a uniform Coordinate-Entropy baseline.
Unlike concurrent GUI confidence estimation methods that sample $K$ stochastic
predictions (SafeGround~\citep{safeground2026}), train an extra
verbalized head (HyperClick~\citep{hyperclick2025}), or reduce the
coordinate distribution to a point statistic (Peak Sharpness
Score~\citep{tao2025localization}), PACE reads only per-token entropies
the model already computes during greedy decoding, so it requires no
training, no architectural change, and no extra forward pass.

We evaluate PACE on ScreenSpot-Pro and ScreenSpot-v$2$~\citep{osatlas2024}
across $7$ open GUI agents, comparing against $12$ single-pass and
sampling-based confidence baselines on three primary agents.
PACE$_{\alpha{=}1}$ wins both AUROC and selective accuracy on all $6$
primary cells in a single forward pass, ahead of $K$-sample baselines
that pay an order of magnitude more in inference, and generally improves uniform
Coordinate-Entropy along the full coverage--accuracy curve. We further study how much of the gain comes from using coordinate-token
  entropy and how much comes from place-value weighting, and discuss a
  quantile-based rule for downstream abstention.

Our contributions are:
\begin{itemize}
\setlength{\itemsep}{1pt}\setlength{\topsep}{2pt}
\item We identify \emph{place-value asymmetry} in coordinate-token sequences and propose \textbf{PACE}, a parameter-free, training-free confidence signal that weights each digit's entropy by its place value.
\item Across fixed-scale agents on two benchmarks, PACE achieves the
  best AUROC and AUARC in all primary comparisons in a single forward
  pass and generally improves uniform Coordinate-Entropy across the
  broader evaluation.
\item We analyze where the gain comes from by comparing full-sequence
  entropy, uniform Coordinate-Entropy, and PACE, separating the benefit
  of using coordinate tokens from the benefit of place-value weighting.
  \end{itemize}

\section{Related Work}
\label{sec:related}

\paragraph{GUI agents and grounding benchmarks.}
Open GUI agents have scaled across multiple generations: from grounding-focused vision-language checkpoints~\citep{cogagent2024,seeclick2024}, through action-supervised trajectory-trained models~\citep{osatlas2024,lin2024showui,aguvis2024,uitars2025}, to recent state-of-the-art checkpoints on professional desktop tasks~\citep{evocua2026,maiui2025}. Closed-API prompted agents~\citep{zheng2024seeact} sit alongside these as a parallel paradigm. Benchmarks have scaled from consumer UIs~\citep{osatlas2024} to professional UIs~\citep{screenspotpro2025} and full-trajectory tasks~\citep{osworld2024,rawles2024androidworld}. Yet evaluation focuses on grounding accuracy or step success; per-prediction confidence remains an open channel, and the Trustworthy GUI Agents survey~\citep{trustworthygui2025} catalogues the safety gap.

\paragraph{GUI-specific confidence estimation.}
Concurrent GUI-specific confidence estimation falls along three lines. Peak Sharpness Score~\citep{tao2025localization} reads a per-token point statistic at coordinate positions, discarding the rest of the distributional shape. SafeGround~\citep{safeground2026} samples $K$ stochastic clicks at temperature $T$ and uses their spatial entropy as the uncertainty signal for the original greedy-decoded click, paying roughly an order of magnitude more in inference for the uncertainty estimate alone. HyperClick~\citep{hyperclick2025} trains a verbalized confidence head with auxiliary reward, requiring supervised retraining. PACE differs from all three: it reads full distribution shape at each coordinate-digit position rather than a point statistic, and runs in a single greedy pass with no sampling, training, or architectural change.

\paragraph{LLM confidence estimation and abstention.}
Text-LLM confidence estimation has matured along several lines: verbalized prompts that ask the model to self-report~\citep{tian2023calibration,kadavath2022language}, sampling agreement over multiple decodes~\citep{wang2023selfconsistency}, self-revision distance~\citep{madaan2023selfrefine}, semantic entropy that clusters samples by meaning~\citep{kuhn2023semantic,farquhar2024semantic}, and post-hoc calibrators applied to model logits~\citep{guo2017calibration,xiong2024llmuncertainty}. A recent abstention survey~\citep{wen2025abstention} catalogues these methods on text-domain tasks, leaving pixel-grounding outside the surveyed scope. Concurrent agentic uncertainty work~\citep{htc2026,auq2026} extends these primitives to trajectory- or task-level aggregation for autonomous agents; PACE supplies the per-action confidence that trajectory-level aggregators consume. PACE can also serve as a cheap upstream gate for sampling-based test-time scaling~\citep{snell2025tts}, deciding which inputs deserve $K$-sample refinement.

\section{Method}
\label{sec:method}

\subsection{GUI Click Confidence Estimation as a Selective-Prediction Task}
\label{sec:task}

A GUI grounding model is a function $f$ that maps a screenshot $I$ and a natural-language instruction $\ell$ to a predicted on-screen click $\hat{y} \in \mathbb{R}^2$ encoded as a digit-token sequence in the model's output vocabulary. The ground-truth target for an instance $(I, \ell)$ is a bounding box $B \subseteq \mathbb{R}^2$, and the binary correctness label is
\begin{equation}
y \;=\; \mathbf{1}\bigl[\hat{y} \in B\bigr].
\label{eq:correctness}
\end{equation}
 A \emph{confidence score} is any scalar $c \colon (I, \ell) \to
  \mathbb{R}$ derived from the model's forward pass that is intended to
  rank predicted clicks by their probability of being correct. We say a
  confidence score is \emph{useful} when clicks with higher $c$ are more
  often correct (discrimination). We evaluate discrimination with AUROC
  and the selective-accuracy integral AUARC. The score $c$ should also be
  cheap to compute, ideally falling out of the same forward pass the model
  already runs to predict $\hat{y}$.

\subsection{Place-Aware Coordinate Entropy (PACE)}
\label{sec:pwce}

Let $z_1, \ldots, z_L$ be the logit vectors at the $L$ coordinate-token positions produced during the model's greedy decode, $p_k = \mathrm{softmax}(z_k)$ the per-position categorical distribution over the vocabulary, and $H(p_k) = -\sum_v p_k(v) \log p_k(v)$ the corresponding Shannon entropy. For an $L$-digit coordinate string, position $k$ carries place value $v_k = 10^{L-1-k}$. We aggregate the per-position entropies as a weighted average,
\begin{equation}
\text{PACE}_\alpha \;=\; -\,\frac{\sum_{k=1}^{L} v_k^{\alpha} \, H(p_k)}{\sum_{k=1}^{L} v_k^{\alpha}}\,,
\label{eq:pwce}
\end{equation}
where the leading negation reorients the score so that larger means more confident. The exponent $\alpha \geq 0$ controls how aggressively place value scales the weights. At $\alpha{=}1$ the hundreds digit carries $100\times$ the weight of the units digit, consistent with the expected pixel deviation of a one-digit error at that place. At $\alpha{=}0$ all positions contribute equally, yielding our Coordinate-Entropy baseline.

The $\alpha{=}1$ choice has a natural pixel-cost interpretation. Under a stylized noise model in which a one-digit confusion at place $k$ shifts the coordinate by $10^{L-1-k}$ pixels, the expected pixel error contributed by entropy at position $k$ scales as $10^{L-1-k} \cdot H(p_k)$. Summing and normalizing recovers Equation~\ref{eq:pwce} with $\alpha{=}1$. Entropy is not literally expected absolute pixel displacement, so this is a heuristic motivation rather than a first-principles derivation; we adopt $\alpha{=}1$ as the simplest parameter-free weighting consistent with place-value asymmetry, and validate empirically (\S\ref{sec:perpos_asymmetry}) that a data-driven linear fit recovers the same ordering.

Implementation: greedy-decode the click, identify coordinate-digit positions in the output, then evaluate Equation~\ref{eq:pwce} from the cached per-step logits. Pseudocode and tokenization edge cases in Appendix~\ref{app:algorithm}.

\section{Experiments}
\label{sec:experiments}

\begin{table*}[!t]
\centering
\footnotesize
\setlength{\tabcolsep}{3pt}
\resizebox{0.90\textwidth}{!}{%
\begin{tabular}{l|cc|cc|cc|c}
\toprule
& \multicolumn{2}{c|}{\textbf{EvoCUA-$8$B}} & \multicolumn{2}{c|}{\textbf{MAI-UI-$8$B}} & \multicolumn{2}{c|}{\textbf{UGround-V$1$-$7$B}} & \\
\textbf{Method} & AUROC$\uparrow$ & AUARC$\uparrow$ & AUROC$\uparrow$ & AUARC$\uparrow$ & AUROC$\uparrow$ & AUARC$\uparrow$ & \textbf{Cost} \\
\midrule
\multicolumn{8}{l}{\emph{Verbalized signals}} \\
Self-Refine             & 0.547 & 0.407 & 0.513 & 0.495 & 0.404 & 0.111 & 2$\times$ \\
Just Ask $1{-}10$       & 0.395 & 0.337 & 0.423 & 0.489 & 0.463 & 0.143 & 1$\times$ \\
P(True)                 & 0.435 & 0.365 & 0.760 & 0.731 & 0.736 & 0.293 & 2$\times$ \\
\midrule
\multicolumn{8}{l}{\emph{Full-sequence signals:}} \\
Logit-min               & 0.726 & 0.641 & 0.643 & 0.669 & 0.761 & 0.419 & 1$\times$ \\
Logit-mean              & 0.764 & 0.675 & 0.723 & 0.710 & 0.816 & 0.455 & 1$\times$ \\
Full-seq entropy        & 0.763 & 0.676 & 0.769 & 0.738 & 0.847 & 0.484 & 1$\times$ \\
\midrule
\multicolumn{8}{l}{\emph{Sampling-based ($K{=}10$):}} \\
 SafeGround IE
  & 0.724 & 0.659
  & 0.673 & 0.671
  & 0.810 & 0.461
  & 10$\times$ \\
 SafeGround combined
  & 0.704 & 0.649
  & 0.683 & 0.679
  & 0.802 & 0.455
  & 10$\times$ \\
\midrule
\multicolumn{8}{l}{\emph{Coordinate-token signals (single forward pass):}} \\
Coord max-entropy       & 0.780 & 0.746 & 0.652 & 0.705 & 0.812 & 0.466 & 1$\times$ \\
PSS top-$2$ gap         & 0.826 & 0.774 & 0.718 & 0.736 & 0.824 & 0.458 & 1$\times$ \\
Coord mean top-$1$ prob & 0.837 & 0.781 & 0.728 & 0.742 & 0.850 & 0.479 & 1$\times$ \\
Coordinate-Entropy  & \underline{0.847} & \underline{0.796} & \underline{0.759} & \underline{0.763} & \underline{0.876} & \underline{0.501} & 1$\times$ \\
\textbf{PACE$_{\alpha{=}1}$ (ours)}  & \textbf{0.857} & \textbf{0.803} & \textbf{0.790} & \textbf{0.797} & \textbf{0.900} & \textbf{0.513} & 1$\times$ \\
\midrule
\multicolumn{8}{l}{\emph{Reference: perfect-ranking ceiling (acc-dependent for AUARC):}} \\
\textit{Perfect ranking}    & 1.000 & 0.879 & 1.000 & 0.912 & 1.000 & 0.592 & --- \\
\bottomrule
\end{tabular}}
\caption{\textbf{Confidence estimation on ScreenSpot-Pro across three primary agents.}
  Bold and underlining mark the best and second-best values in each column.
  All methods, including sampling-based methods, score the same greedy-decoded click. Paired results on ScreenSpot-v$2$ are reported in Appendix.}
\label{tab:headline}
\end{table*}

\subsection{Setup}

\paragraph{Datasets.}
ScreenSpot-Pro~\citep{screenspotpro2025} provides $1{,}581$ click-grounding tasks across $23$ professional desktop applications such as Photoshop, AutoCAD, and MATLAB. ScreenSpot-v$2$~\citep{osatlas2024} provides $1{,}272$ click-grounding tasks across consumer mobile, web, and desktop UIs. Each task is an (instruction, screenshot, target bounding box) triple; the agent's predicted click is scored correct if it falls inside the target box.

\paragraph{Models.}
The three primary agents are EvoCUA-$8$B~\citep{evocua2026}, MAI-UI-$8$B~\citep{maiui2025}, and UGround-V$1$-$7$B~\citep{uground2025}. Four additional agents (UGround-V$1$-$2$B, UI-TARS-$7$B-SFT, UI-TARS-$7$B-DPO~\citep{uitars2025}, Aguvis-$7$B-720P~\citep{aguvis2024}) support the cross-agent paired analysis. 

\paragraph{Confidence estimation baselines.}
We compare PACE against $12$ baselines in four categories, ordered to match Table~\ref{tab:headline}.
\textbf{Verbalized} signals query the model directly: Self-Refine~\citep{madaan2023selfrefine} runs one round of self-critique before re-prompting for confidence; Just Ask $1$--$10$~\citep{tian2023calibration} returns an integer on a $1$ to $10$ scale; P(True)~\citep{kadavath2022language} asks an A/B correctness question and reads the softmax probability of ``A''.
\textbf{Full-sequence} signals (logit-min, logit-mean, full-sequence Shannon entropy) aggregate over every output token, coordinate-digit and surface-form alike.
\textbf{Sampling-based} signals draw $K{=}10$ samples at temperature $T{=}1.0$: SafeGround entropy is the density entropy over the $K$ predicted click positions, and SafeGround combined~\citep{safeground2026} adds patch-level disagreement on top.
\textbf{Coordinate-token} signals read only the coordinate-digit positions: coordinate max-entropy, the worst-position entropy; PSS~\citep{tao2025localization}, the top-$1$/top-$2$ logit margin at each position; coordinate mean top-$1$ probability (the average greedy-token softmax probability across coordinate positions); and Coordinate-Entropy (PACE$_{\alpha{=}0}$, the uniform-weight baseline).

\paragraph{Metrics.}
  AUROC ranks correct against incorrect clicks by confidence score.
  AUARC is the selective-accuracy integral: how much accuracy the agent
  retains when allowed to abstain on its lowest-confidence fraction. All
  signals are evaluated on the same greedy clicks; sampling-based methods
  use their samples only to estimate uncertainty.

\subsection{Main Results}
\label{sec:main_result}

\paragraph{Comparisons with baselines.}
Table~\ref{tab:headline} compares PACE with twelve confidence baselines from four families across the three primary agents. PACE$_{\alpha{=}1}$ achieves the highest
  AUROC and AUARC on all three agents using a single forward pass, outperforming sampling baselines that use $K{=}10$ stochastic decodes per decision. Across the
  three agents, PACE reaches $87$--$91\%$ of the perfect-ranking AUARC ceiling, leaving a relatively small absolute gap to the oracle ranking on this metric.

\paragraph{Decomposing the gain.}
  The PACE$_{\alpha{=}1}$ advantage comes from two components. First, Shannon entropy at coordinate-digit positions alone (Coordinate-Entropy) already outperforms the
  strongest alternative single-pass summaries by $0.02$--$0.08$ mean AUROC across the three primary agents. The reason is distributional: these summaries retain only
  part of the predictive distribution. PSS uses only the top-two logit gap, the mean top-one probability uses only the greedy-token probability at each coordinate
  position, and max-entropy uses only the worst position. When uncertainty is spread across several digit candidates, these summaries can miss part of the
  distributional structure; Shannon entropy instead uses the full categorical distribution at each position. Second, place-value weighting adds $0.022$ mean AUROC
  over Coordinate-Entropy, with positive gains on all three primary agents. 

 \paragraph{Response-level confidence can differ from click accuracy.}
  The ranking metrics above measure whether a confidence score orders
  correct and incorrect clicks. We separately examine how the standard
  logit-mean score relates to task accuracy, where Conf is the per-token
  top-1 softmax probability averaged across the response. For MAI-UI-8B
  and EvoCUA-8B, logit-mean confidence remains high on ScreenSpot-Pro
  even though click accuracy is much lower. On ScreenSpot-v2, both
  confidence and accuracy are high (Table~\ref{tab:main}).
  These results show that averaging confidence over the full response
  can hide uncertainty concentrated in the coordinate digits that
  determine the click. This motivates a coordinate-local signal. PACE
  reads entropy at those positions and is evaluated as a ranking signal
  rather than a calibrated probability.

\begin{table}[H]
  \centering
  \small
  \setlength{\tabcolsep}{4pt}
  \resizebox{\columnwidth}{!}{%
  \begin{tabular}{l|ccc|ccc}
  \toprule
  & \multicolumn{3}{c|}{\textbf{ScreenSpot-Pro}}
  & \multicolumn{3}{c}{\textbf{ScreenSpot-v$2$}} \\
  \textbf{Model} & Acc & Conf & Gap & Acc & Conf & Gap \\
  \midrule
  MAI-UI-$8$B & $0.494$ & $0.939$ & $0.445$
               & $0.945$ & $0.952$ & $0.007$ \\
  EvoCUA-$8$B & $0.352$ & $0.833$ & $0.481$
               & $0.910$ & $0.875$ & $-0.035$ \\
  \bottomrule
  \end{tabular}}

  \caption{\textbf{Response-level token confidence and click accuracy.}
  Conf is the logit-mean, computed as the per-token top-$1$ softmax
  probability averaged across the response; Gap ${=}$ Conf ${-}$ Acc.}
  \label{tab:main}
  \end{table}

  \subsection{Cross-Agent Performance}

  The place-weighting gain extends beyond the three primary agents to a broader eight-agent pool on both benchmarks (Table~\ref{tab:pwce_universal_6model}).
  PACE$_{\alpha{=}1}$ improves over Coordinate-Entropy for all eight agents on both ScreenSpot-Pro and ScreenSpot-v$2$. The mean gain is noticeably larger on consumer
  ScreenSpot-v$2$ than on professional ScreenSpot-Pro despite v$2$'s higher base accuracy. 

\begin{table}[!htb]
  \centering
  \small
  \setlength{\tabcolsep}{4pt}
  \resizebox{\columnwidth}{!}{%
  \begin{tabular}{l|cc|cc}
  \toprule
  & \multicolumn{2}{c|}{\textbf{ScreenSpot-Pro}}
  & \multicolumn{2}{c}{\textbf{ScreenSpot-v$2$}} \\
  \textbf{Model}
  & $\alpha{=}0$ & $\alpha{=}1$
  & $\alpha{=}0$ & $\alpha{=}1$ \\
  \midrule
  UI-TARS-$7$B-DPO
  & $0.777$ & $\mathbf{0.785}$
  & $0.654$ & $\mathbf{0.762}$ \\

  MAI-UI-$8$B
  & $0.759$ & $\mathbf{0.790}$
  & $0.580$ & $\mathbf{0.596}$ \\

  UI-TARS-$7$B-SFT
  & $0.807$ & $\mathbf{0.822}$
  & $0.678$ & $\mathbf{0.770}$ \\

  UI-TARS-$72$B-DPO
  & $0.808$ & $\mathbf{0.833}$
  & $0.637$ & $\mathbf{0.738}$ \\

  Aguvis-$7$B-$720$P
  & $0.836$ & $\mathbf{0.837}$
  & $0.699$ & $\mathbf{0.728}$ \\

  EvoCUA-$8$B
  & $0.847$ & $\mathbf{0.857}$
  & $0.738$ & $\mathbf{0.803}$ \\

  UGround-V$1$-$2$B
  & $0.854$ & $\mathbf{0.890}$
  & $0.773$ & $\mathbf{0.837}$ \\

  UGround-V$1$-$7$B
  & $0.876$ & $\mathbf{0.900}$
  & $0.723$ & $\mathbf{0.777}$ \\
  \midrule
 \textbf{Mean $\Delta$}
  & \multicolumn{2}{c|}{$\mathbf{+0.019}$}
  & \multicolumn{2}{c}{$\mathbf{+0.066}$} \\
  \bottomrule
  \end{tabular}}
  \caption{\textbf{Cross-agent comparison of PACE$_{\alpha{=}1}$ and Coordinate-Entropy.}}
  \label{tab:pwce_universal_6model}
  \end{table}


\section{Analysis}
\label{sec:analysis}

\subsection{Why PACE Works: Hundreds Digit Carries the Signal}
\label{sec:mechanistic}

\begin{figure}[!htb]
\centering
\includegraphics[width=\columnwidth]{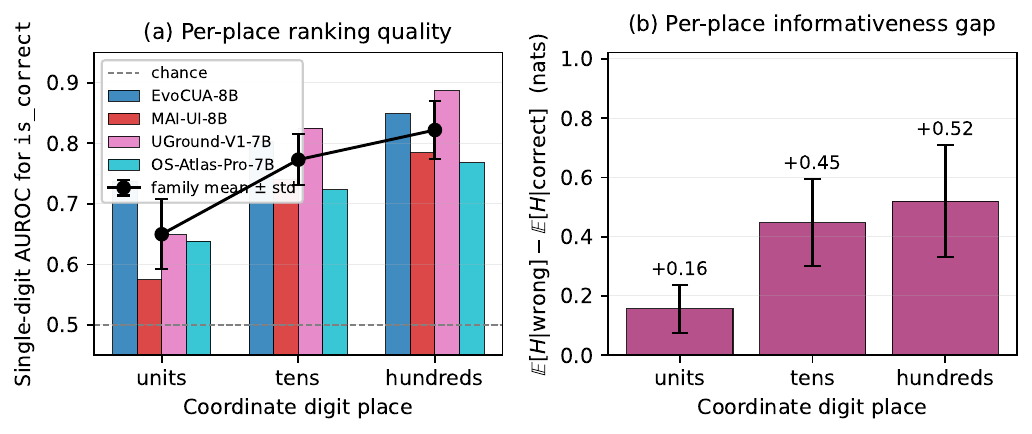}
\caption{\textbf{Per-place informativeness justifies place-weighting.} (a) Single-digit AUROC at each digit place; family mean rises monotonically units to hundreds. (b) Per-place entropy gap between wrong and correct predictions tracks the same gradient.}
\label{fig:perplace_just}
\end{figure}

A one-digit confusion at the hundreds place (for example $x{=}502$ vs.\ $x{=}602$) changes the normalized coordinate by $100$ grid units and can move the click onto a
  neighboring UI element. The same confusion at the units place ($x{=}502$ vs.\ $x{=}508$) changes the coordinate by only $6$ grid units and is more likely to remain
  inside the target. Per-position uncertainty is therefore not interchangeable: hundreds-digit uncertainty is more predictive of incorrect clicks than units-digit
  uncertainty. When averaged across agents, single-digit entropy AUROC increases from fine to coarse places (Figure~\ref{fig:perplace_just}\textbf{a}), from $0.65$ at
  units to $0.77$ at tens and $0.82$ at hundreds. The expected entropy gap $\mathbb{E}[H \mid \mathrm{wrong}] - \mathbb{E}[H \mid \mathrm{correct}]$ follows the same
  pattern (Figure~\ref{fig:perplace_just}\textbf{b}). PACE$_{\alpha{=}1}$ assigns roughly $90\%$ of its positional weight to the hundreds digit, where the empirical
  signal is strongest. 

\subsection{Why Not Use Only the Hundreds Digit?}
  \label{sec:perpos_asymmetry}

  The natural simplification is to score clicks using hundreds-place entropy alone. Table~\ref{tab:weight_schemes} compares this shortcut with learned linear weights
  and PACE$_{\alpha{=}1}$ using mean AUROC across the three primary agents.

  \begin{table}[!htb]
  \centering
  \small
  \setlength{\tabcolsep}{4pt}
  \resizebox{\columnwidth}{!}{%
  \begin{tabular}{l|c|c}
  \toprule
  \textbf{Scheme}
  & \textbf{Normalized weights $(v_h,v_t,v_u)$}
  & \textbf{Mean AUROC} \\
  \midrule
  Hundreds only
  & $(1.000,0.000,0.000)$
  & $0.847$ \\
  LR-learned (pooled, $\ell_1$)
  & $(0.710,0.280,0.010)$
  & $\mathbf{0.851}$ \\
  \textbf{PACE$_{\alpha{=}1}$}
  & $(0.901,0.090,0.009)$
  & $\mathbf{0.851}$ \\
  \bottomrule
  \end{tabular}}
  \caption{\textbf{Hundreds-only and learned weighting compared with PACE$_{\alpha{=}1}$.} LR weights are
  learned on pooled data with $\ell_1$ regularization and $5$-fold cross-validation.}
  \label{tab:weight_schemes}
  \end{table}

   Hundreds-only is close to PACE on the aggregate evaluation, but this does not make the lower places dispensable. When the predicted click and target center fall in
  the same hundreds cell, correctness depends more directly on lower-place precision. Table~\ref{tab:same_hundreds} shows this effect for EvoCUA-$8$B.

  \begin{table}[!htb]
  \centering
  \small
  \setlength{\tabcolsep}{8pt}
  \begin{tabular}{lc}
  \toprule
  \textbf{Digit} & \textbf{Standalone AUROC} \\
  \midrule
  Hundreds & $.746$ \\
  Tens     & $\mathbf{.812}$ \\
  Units    & $.801$ \\
  \bottomrule
  \end{tabular}
  \caption{\textbf{Standalone digit discrimination within the same hundreds cell.} Results are for EvoCUA-$8$B on ScreenSpot-Pro examples where the predicted click
  and target center share the same hundreds cell ($n{=}619$).}
  \label{tab:same_hundreds}
  \end{table}

  The tens digit becomes the most informative individual place in this subset. This shift is not specific to EvoCUA: tens is the strongest digit in $11$ of the $14$
  evaluated same-hundreds settings. These results show that lower-place entropy is not redundant even when the hundreds digit dominates aggregate performance. A
  hundreds-only score can discard this fine-place information, which motivates retaining the full place-value formula in PACE.

  The pooled linear fit recovers a hundreds-dominant profile and matches PACE's mean AUROC. This suggests that performance is not sensitive to the exact weight ratio
  within this ordering. PACE obtains its weights directly from decimal place value without fitting. A fine-grid sweep remains nearly flat for $\alpha \in [0.3,2.0]$,
  while shuffling the place-value assignment reduces mean AUROC by more than four points ($p{=}0.04$; Appendix~\ref{app:alpha_finegrid}). Together, these results support
  $\alpha{=}1$ as a robust default while showing that the assignment of weights to digit places matters.

\subsection{Visualizing Signal Quality}
\label{sec:spatial_failure}

\begin{figure*}[!t]
\centering
\includegraphics[width=\textwidth]{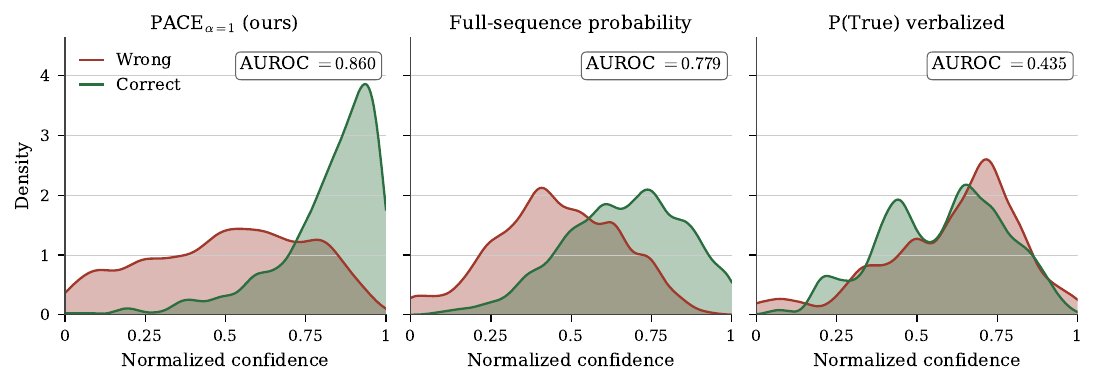}
\caption{\textbf{Confidence distributions on EvoCUA-$8$B ScreenSpot-Pro.} PACE$_{\alpha{=}1}$ shows the clearest separation between correct (green) and incorrect
  (red) clicks. Full-sequence probability provides partial separation, while verbalized P(True) produces substantially overlapping distributions. AUROC values are shown
  within each panel.}
\label{fig:signal_distribution}
\end{figure*}

Figure~\ref{fig:signal_distribution} contrasts the EvoCUA-$8$B confidence distributions of three signal families on ScreenSpot-Pro: PACE$_{\alpha{=}1}$, confidence
  derived from full-sequence probability, and verbalized P(True). PACE shows the clearest separation between correct (green) and incorrect (red) clicks, while full-
  sequence probability provides only partial separation. The P(True) distributions overlap substantially, consistent with its below-chance AUROC in this setting. These
  distributions help explain the selective-prediction results: PACE provides a more useful confidence threshold for deciding whether to execute a click or hand it
  off, whereas P(True) offers little separation for that decision. Appendix~\ref{app:structural_properties} analyzes the corresponding limitations of full-sequence
  aggregation, verbalized confidence, and self-consistency.

\subsection{Downstream: Selective Prediction and Abstention}
\label{sec:deployment}

A deployed GUI agent does not act on a confidence ranking metric directly. It must decide for each click whether to execute it or abstain. Selective
  prediction~\citep{geifman2017selective} evaluates this decision through the accuracy of executed clicks at different coverage levels. On the five-agent mean,
  PACE$_{\alpha{=}1}$ outperforms Coordinate-Entropy at every coverage level reported in Table~\ref{tab:sel_acc_pwce}. The difference is largest at low coverage,
  where the agent executes only its highest-confidence clicks and abstains on the rest. At $5\%$ coverage, PACE raises mean selective accuracy from $63.8\%$ to $73.2\%$. The advantage becomes smaller as coverage increases, indicating that place weighting is most useful for identifying the highest-confidence subset.

\begin{table}[!htb]
\centering
\small
\setlength{\tabcolsep}{3pt}
\resizebox{\columnwidth}{!}{%
\begin{tabular}{l|cc|cc|cc|cc}
\toprule
& \multicolumn{2}{c|}{\textbf{5\%}} & \multicolumn{2}{c|}{\textbf{10\%}} & \multicolumn{2}{c|}{\textbf{20\%}} & \multicolumn{2}{c}{\textbf{50\%}} \\
\textbf{Agent} & $\alpha{=}0$ & $\alpha{=}1$ & $\alpha{=}0$ & $\alpha{=}1$ & $\alpha{=}0$ & $\alpha{=}1$ & $\alpha{=}0$ & $\alpha{=}1$ \\
\midrule
EvoCUA-$8$B       & $97.7$ & $97.7$            & $97.7$ & $97.7$            & $93.8$ & $93.2$            & $80.3$ & $\mathbf{81.0}$ \\
MAI-UI-$8$B       & $78.0$ & $\mathbf{90.0}$   & $81.0$ & $\mathbf{88.0}$   & $81.0$ & $\mathbf{88.5}$   & $80.0$ & $\mathbf{82.8}$ \\
UGround-V$1$-$7$B & $87.1$ & $\mathbf{91.9}$   & $83.7$ & $82.1$            & $71.5$ & $71.5$            & $44.7$ & $\mathbf{46.7}$ \\
UI-TARS-$7$B-SFT  & $32.4$ & $\mathbf{46.0}$   & $50.7$ & $\mathbf{56.2}$   & $48.6$ & $\mathbf{50.0}$   & $34.4$ & $\mathbf{35.3}$ \\
UI-TARS-$7$B-DPO  & $23.8$ & $\mathbf{40.5}$   & $51.8$ & $\mathbf{55.4}$   & $63.5$ & $63.5$            & $49.5$ & $\mathbf{49.8}$ \\
\midrule
\textbf{Mean}     & $63.8$ & $\mathbf{73.2}$   & $73.0$ & $\mathbf{75.9}$   & $71.7$ & $\mathbf{73.3}$   & $57.8$ & $\mathbf{59.1}$ \\
\bottomrule
\end{tabular}}

 \caption{\textbf{Selective accuracy at fixed coverage levels.} PACE$_{\alpha{=}1}$ improves the five-agent mean over Coordinate-Entropy ($\alpha{=}0$) at every
  reported level. Bold marks individual settings in which PACE obtains higher selective accuracy; ties and lower PACE values are retained.}
\label{tab:sel_acc_pwce}
\end{table}

 \paragraph{Wrong-click prevention.}
Table~\ref{tab:wrong_click_prevention} reports representative abstention settings for EvoCUA-$8$B. PACE$_{\alpha{=}1}$ removes at least as many wrong clicks as
  Coordinate-Entropy on ScreenSpot-Pro and substantially more on ScreenSpot-v$2$.  These results complement selective accuracy by
  showing directly how often confidence-based abstention prevents an incorrect click from being executed.

  \begin{table}[!htb]
  \centering
  \small
  \setlength{\tabcolsep}{5pt}
  \begin{tabular}{lccc}
  \toprule
  \textbf{Benchmark}
  & \textbf{Abstention rate}
  & \textbf{PACE}
  & \textbf{CE} \\
  \midrule
  ScreenSpot-Pro
  & $30\%$ & $\mathbf{56.1\%}$ & $55.9\%$ \\
  ScreenSpot-Pro
  & $50\%$ & $\mathbf{78.6\%}$ & $77.9\%$ \\
  ScreenSpot-v$2$
  & $10\%$ & $\mathbf{36.7\%}$ & $29.1\%$ \\
  ScreenSpot-v$2$
  & $30\%$ & $\mathbf{72.2\%}$ & $59.5\%$ \\
  \bottomrule
  \end{tabular}
  \caption{\textbf{Wrong-click prevention for EvoCUA-$8$B.} Values report the percentage of incorrect clicks not executed at each abstention rate. PACE denotes
  PACE$_{\alpha{=}1}$, and CE denotes Coordinate-Entropy.}
  \label{tab:wrong_click_prevention}
  \end{table}
  \paragraph{Cross-agent routing.}
  PACE can also select among candidate clicks from multiple agents. For each example, we execute the click proposed by the most confident agent in a six-agent pool.
  Table~\ref{tab:cross_agent_routing} compares routing by PACE and Coordinate-Entropy with always executing the strongest fixed agent.

  \begin{table}[!htb]
  \centering
  \small
  \setlength{\tabcolsep}{8pt}
  \begin{tabular}{lc}
  \toprule
  \textbf{Selection rule} & \textbf{Accuracy} \\
  \midrule
  \textbf{PACE$_{\alpha{=}1}$ routing} & $\mathbf{.717}$ \\
  Always-best single agent             & $.711$ \\
  Coordinate-Entropy routing           & $.693$ \\
  \bottomrule
  \end{tabular}
  \caption{\textbf{Cross-agent routing on ScreenSpot-Pro.} Each method selects one click from a common six-agent pool of $512$ examples. PACE improves over
  Coordinate-Entropy routing, while its difference from the strongest fixed agent is small.}
  \label{tab:cross_agent_routing}
  \end{table}

  PACE routing improves accuracy over Coordinate-Entropy routing by $2.3$ percentage points. Its smaller advantage over the strongest fixed agent is within
  statistical noise and should not be interpreted as a reliable improvement. In the three-agent pool, PACE and Coordinate-Entropy make the same routing decisions,
  indicating that the benefit appears when the candidate pool is larger and more heterogeneous.

\section{Conclusion}
\label{sec:conclusion}

  GUI agents often emit clicks as short digit-token sequences on a fixed coordinate grid. Standard text-LLM confidence methods aggregate across tokens or coordinate
  positions without accounting for place value asymmetry. Place-Aware Coordinate Entropy addresses this mismatch by computing Shannon entropy at coordinate-
  token positions and weighting each digit by its decimal place value. The resulting confidence score requires only the logits from a single greedy decode. Across the
  three primary agents, PACE$_{\alpha{=}1}$ achieves the best AUROC and AUARC in all six agent and metric comparisons. In the eight-agent comparison, place weighting
  improves mean AUROC on both ScreenSpot-Pro and ScreenSpot-v$2$, including a positive result for a $72$B fixed-scale model.

  PACE also supports a direct execution decision: the agent executes high-confidence clicks and abstains on low-confidence ones. Across the evaluated fixed-scale
  agents, PACE prevents more wrong clicks than Coordinate-Entropy at most reported abstention settings, with its largest gains appearing at low coverage. These
  results connect the ranking improvement to fewer incorrect clicks being executed. The evidence is scoped to coordinate formats with separately tokenized digits and
  a stable place-value map. For variable-scale pixel coordinates, fractional outputs, or merged digit tokens, decimal place weighting is not guaranteed to help, and
  Coordinate-Entropy remains the appropriate fallback. One possible future work is to explore how PACE can guide abstention and recovery decisions in closed-loop, multi-step GUI agents.

 \section*{Limitations}

  \begin{itemize}[leftmargin=*,topsep=2pt,itemsep=2pt,parsep=0pt]

  \item \textbf{Closed-loop interaction.}
  Our main evaluation focuses on individual grounding decisions on ScreenSpot-Pro and ScreenSpot-v$2$. It does not establish how confidence-based abstention affects
  long-horizon task completion. Future work should evaluate PACE in closed-loop, multi-step GUI agents.

  \item \textbf{Emission format.}
  PACE assumes that coordinates are emitted as separately tokenized digits on a stable scale. Decimal place weighting is not guaranteed to help with variable-scale
  pixel coordinates, fractional outputs, or merged digit tokens; Coordinate-Entropy is the appropriate fallback in these settings.

  \item \textbf{Model coverage.}
  Although our evaluation spans models from $2$B to $72$B, the large-scale evidence comes from a single $72$B fixed-scale agent. Additional large and proprietary
  agents are needed to separate the effects of model scale, architecture, and coordinate format.

  \item \textbf{Discrimination, not probability calibration.}
  PACE is designed to rank clicks by confidence rather than estimate their probability of correctness. It can be thresholded directly for abstention, but applications
  that require calibrated error probabilities need a separate post-hoc calibration procedure.

  \end{itemize}

\section*{Ethical Considerations}

GUI agents operate on interfaces whose actions are not always reversible~\citep{trustworthygui2025}; the cost of a misgrounded click extends beyond the immediate task. PACE is a per-step uncertainty primitive for downstream safeguards (clarification, human review, abstention, escalation), not a standalone deployment certificate. Safety-critical deployment should include per-application monitoring, recomputation of the calibration set under distribution shift, and clear allocation of responsibility between the automated decision and human oversight.

\section*{Acknowledgments}

This work is supported by Centre for Perceptual and Interactive Intelligence (CPII) Ltd, a
CUHK-led InnoCentre under InnoHK scheme of Innovation and Technology Commission.

\bibliography{references}

@article{evocua2026,
  author       = {Taofeng Xue and
                  Chong Peng and
                  Mianqiu Huang and
                  Linsen Guo and
                  Tiancheng Han and
                  Haozhe Wang and
                  Jianing Wang and
                  Xiaocheng Zhang and
                  Xin Yang and
                  Dengchang Zhao and
                  Jinrui Ding and
                  Xiandi Ma and
                  Yuchen Xie and
                  Peng Pei and
                  Xunliang Cai and
                  Xipeng Qiu},
  title        = {EvoCUA: Evolving Computer Use Agents via Learning from Scalable Synthetic
                  Experience},
  journal      = {CoRR},
  volume       = {abs/2601.15876},
  year         = {2026},
  url          = {https://doi.org/10.48550/arXiv.2601.15876},
  doi          = {10.48550/ARXIV.2601.15876},
  eprinttype   = {arXiv},
  eprint       = {2601.15876},
  bibsource    = {dblp computer science bibliography, https://dblp.org}
}

@article{uitars2025,
  author       = {Yujia Qin and
                  Yining Ye and
                  Junjie Fang and
                  Haoming Wang and
                  Shihao Liang and
                  Shizuo Tian and
                  Junda Zhang and
                  Jiahao Li and
                  Yunxin Li and
                  Shijue Huang and
                  Wanjun Zhong and
                  Kuanye Li and
                  Jiale Yang and
                  Yu Miao and
                  Woyu Lin and
                  Longxiang Liu and
                  Xu Jiang and
                  Qianli Ma and
                  Jingyu Li and
                  Xiaojun Xiao and
                  Kai Cai and
                  Chuang Li and
                  Yaowei Zheng and
                  Chaolin Jin and
                  Chen Li and
                  Xiao Zhou and
                  Minchao Wang and
                  Haoli Chen and
                  Zhaojian Li and
                  Haihua Yang and
                  Haifeng Liu and
                  Feng Lin and
                  Tao Peng and
                  Xin Liu and
                  Guang Shi},
  title        = {{UI-TARS:} Pioneering Automated {GUI} Interaction with Native Agents},
  journal      = {CoRR},
  volume       = {abs/2501.12326},
  year         = {2025},
  url          = {https://doi.org/10.48550/arXiv.2501.12326},
  doi          = {10.48550/ARXIV.2501.12326},
  eprinttype   = {arXiv},
  eprint       = {2501.12326},
  bibsource    = {dblp computer science bibliography, https://dblp.org}
}

@inproceedings{cogagent2024,
  author       = {Wenyi Hong and
                  Weihan Wang and
                  Qingsong Lv and
                  Jiazheng Xu and
                  Wenmeng Yu and
                  Junhui Ji and
                  Yan Wang and
                  Zihan Wang and
                  Yuxiao Dong and
                  Ming Ding and
                  Jie Tang},
  title        = {CogAgent: {A} Visual Language Model for {GUI} Agents},
  booktitle    = {{IEEE/CVF} Conference on Computer Vision and Pattern Recognition,
                  {CVPR} 2024, Seattle, WA, USA, June 16-22, 2024},
  pages        = {14281--14290},
  publisher    = {{IEEE}},
  year         = {2024},
  url          = {https://doi.org/10.1109/CVPR52733.2024.01354},
  doi          = {10.1109/CVPR52733.2024.01354},
  bibsource    = {dblp computer science bibliography, https://dblp.org}
}

@inproceedings{osworld2024,
  author       = {Tianbao Xie and
                  Danyang Zhang and
                  Jixuan Chen and
                  Xiaochuan Li and
                  Siheng Zhao and
                  Ruisheng Cao and
                  Toh Jing Hua and
                  Zhoujun Cheng and
                  Dongchan Shin and
                  Fangyu Lei and
                  Yitao Liu and
                  Yiheng Xu and
                  Shuyan Zhou and
                  Silvio Savarese and
                  Caiming Xiong and
                  Victor Zhong and
                  Tao Yu},
  editor       = {Amir Globersons and
                  Lester Mackey and
                  Danielle Belgrave and
                  Angela Fan and
                  Ulrich Paquet and
                  Jakub M. Tomczak and
                  Cheng Zhang},
  title        = {OSWorld: Benchmarking Multimodal Agents for Open-Ended Tasks in Real
                  Computer Environments},
  booktitle    = {Advances in Neural Information Processing Systems 38: Annual Conference
                  on Neural Information Processing Systems 2024, NeurIPS 2024, Vancouver,
                  BC, Canada, December 10 - 15, 2024},
  year         = {2024},
  url          = {http://papers.nips.cc/paper\_files/paper/2024/hash/5d413e48f84dc61244b6be550f1cd8f5-Abstract-Datasets\_and\_Benchmarks\_Track.html},
  bibsource    = {dblp computer science bibliography, https://dblp.org}
}

@inproceedings{screenspotpro2025,
  author       = {Kaixin Li and
                  Ziyang Meng and
                  Hongzhan Lin and
                  Ziyang Luo and
                  Yuchen Tian and
                  Jing Ma and
                  Zhiyong Huang and
                  Tat{-}Seng Chua},
  editor       = {Cathal Gurrin and
                  Klaus Schoeffmann and
                  Min Zhang and
                  Luca Rossetto and
                  Stevan Rudinac and
                  Duc{-}Tien Dang{-}Nguyen and
                  Wen{-}Huang Cheng and
                  Phoebe Chen and
                  Jenny Benois{-}Pineau},
  title        = {ScreenSpot-Pro: {GUI} Grounding for Professional High-Resolution Computer
                  Use},
  booktitle    = {Proceedings of the 33rd {ACM} International Conference on Multimedia,
                  {MM} 2025, Dublin, Ireland, October 27-31, 2025},
  pages        = {8778--8786},
  publisher    = {{ACM}},
  year         = {2025},
  url          = {https://doi.org/10.1145/3746027.3755688},
  doi          = {10.1145/3746027.3755688},
  bibsource    = {dblp computer science bibliography, https://dblp.org}
}

@article{maiui2025,
  author       = {Hanzhang Zhou and
                  Xu Zhang and
                  Panrong Tong and
                  Jianan Zhang and
                  Liangyu Chen and
                  Quyu Kong and
                  Chenglin Cai and
                  Chen Liu and
                  Yue Wang and
                  Jingren Zhou and
                  Steven Hoi},
  title        = {{MAI-UI} Technical Report: Real-World Centric Foundation {GUI} Agents},
  journal      = {CoRR},
  volume       = {abs/2512.22047},
  year         = {2025},
  url          = {https://doi.org/10.48550/arXiv.2512.22047},
  doi          = {10.48550/ARXIV.2512.22047},
  eprinttype   = {arXiv},
  eprint       = {2512.22047},
  bibsource    = {dblp computer science bibliography, https://dblp.org}
}

@inproceedings{wang2023selfconsistency,
  author       = {Xuezhi Wang and
                  Jason Wei and
                  Dale Schuurmans and
                  Quoc V. Le and
                  Ed H. Chi and
                  Sharan Narang and
                  Aakanksha Chowdhery and
                  Denny Zhou},
  title        = {Self-Consistency Improves Chain of Thought Reasoning in Language Models},
  booktitle    = {The Eleventh International Conference on Learning Representations,
                  {ICLR} 2023, Kigali, Rwanda, May 1-5, 2023},
  publisher    = {OpenReview.net},
  year         = {2023},
  url          = {https://openreview.net/forum?id=1PL1NIMMrw},
  bibsource    = {dblp computer science bibliography, https://dblp.org}
}

@inproceedings{tian2023calibration,
  author       = {Katherine Tian and
                  Eric Mitchell and
                  Allan Zhou and
                  Archit Sharma and
                  Rafael Rafailov and
                  Huaxiu Yao and
                  Chelsea Finn and
                  Christopher D. Manning},
  editor       = {Houda Bouamor and
                  Juan Pino and
                  Kalika Bali},
  title        = {Just Ask for Calibration: Strategies for Eliciting Calibrated Confidence
                  Scores from Language Models Fine-Tuned with Human Feedback},
  booktitle    = {Proceedings of the 2023 Conference on Empirical Methods in Natural
                  Language Processing, {EMNLP} 2023, Singapore, December 6-10, 2023},
  pages        = {5433--5442},
  publisher    = {Association for Computational Linguistics},
  year         = {2023},
  url          = {https://doi.org/10.18653/v1/2023.emnlp-main.330},
  doi          = {10.18653/V1/2023.EMNLP-MAIN.330},
  bibsource    = {dblp computer science bibliography, https://dblp.org}
}

@article{kadavath2022language,
  author       = {Saurav Kadavath and
                  Tom Conerly and
                  Amanda Askell and
                  Tom Henighan and
                  Dawn Drain and
                  Ethan Perez and
                  Nicholas Schiefer and
                  Zac Hatfield{-}Dodds and
                  Nova DasSarma and
                  Eli Tran{-}Johnson and
                  Scott Johnston and
                  Sheer El Showk and
                  Andy Jones and
                  Nelson Elhage and
                  Tristan Hume and
                  Anna Chen and
                  Yuntao Bai and
                  Sam Bowman and
                  Stanislav Fort and
                  Deep Ganguli and
                  Danny Hernandez and
                  Josh Jacobson and
                  Jackson Kernion and
                  Shauna Kravec and
                  Liane Lovitt and
                  Kamal Ndousse and
                  Catherine Olsson and
                  Sam Ringer and
                  Dario Amodei and
                  Tom Brown and
                  Jack Clark and
                  Nicholas Joseph and
                  Ben Mann and
                  Sam McCandlish and
                  Chris Olah and
                  Jared Kaplan},
  title        = {Language Models (Mostly) Know What They Know},
  journal      = {CoRR},
  volume       = {abs/2207.05221},
  year         = {2022},
  url          = {https://doi.org/10.48550/arXiv.2207.05221},
  doi          = {10.48550/ARXIV.2207.05221},
  eprinttype   = {arXiv},
  eprint       = {2207.05221},
  bibsource    = {dblp computer science bibliography, https://dblp.org}
}

@inproceedings{kuhn2023semantic,
  author       = {Lorenz Kuhn and
                  Yarin Gal and
                  Sebastian Farquhar},
  title        = {Semantic Uncertainty: Linguistic Invariances for Uncertainty Estimation
                  in Natural Language Generation},
  booktitle    = {The Eleventh International Conference on Learning Representations,
                  {ICLR} 2023, Kigali, Rwanda, May 1-5, 2023},
  publisher    = {OpenReview.net},
  year         = {2023},
  url          = {https://openreview.net/forum?id=VD-AYtP0dve},
  bibsource    = {dblp computer science bibliography, https://dblp.org}
}

@inproceedings{guo2017calibration,
  author       = {Chuan Guo and
                  Geoff Pleiss and
                  Yu Sun and
                  Kilian Q. Weinberger},
  editor       = {Doina Precup and
                  Yee Whye Teh},
  title        = {On Calibration of Modern Neural Networks},
  booktitle    = {Proceedings of the 34th International Conference on Machine Learning,
                  {ICML} 2017, Sydney, NSW, Australia, 6-11 August 2017},
  series       = {Proceedings of Machine Learning Research},
  pages        = {1321--1330},
  publisher    = {{PMLR}},
  year         = {2017},
  url          = {http://proceedings.mlr.press/v70/guo17a.html},
  bibsource    = {dblp computer science bibliography, https://dblp.org}
}

@inproceedings{xiong2024llmuncertainty,
  author       = {Miao Xiong and
                  Zhiyuan Hu and
                  Xinyang Lu and
                  Yifei Li and
                  Jie Fu and
                  Junxian He and
                  Bryan Hooi},
  title        = {Can LLMs Express Their Uncertainty? An Empirical Evaluation of Confidence
                  Elicitation in LLMs},
  booktitle    = {The Twelfth International Conference on Learning Representations,
                  {ICLR} 2024, Vienna, Austria, May 7-11, 2024},
  publisher    = {OpenReview.net},
  year         = {2024},
  url          = {https://openreview.net/forum?id=gjeQKFxFpZ},
  bibsource    = {dblp computer science bibliography, https://dblp.org}
}

@article{htc2026,
  author       = {Jiaxin Zhang and
                  Caiming Xiong and
                  Chien{-}Sheng Wu},
  title        = {Agentic Confidence Calibration},
  journal      = {CoRR},
  volume       = {abs/2601.15778},
  year         = {2026},
  url          = {https://doi.org/10.48550/arXiv.2601.15778},
  doi          = {10.48550/ARXIV.2601.15778},
  eprinttype   = {arXiv},
  eprint       = {2601.15778},
  bibsource    = {dblp computer science bibliography, https://dblp.org}
}

@article{auq2026,
  author       = {Jiaxin Zhang and
                  Prafulla Kumar Choubey and
                  Kung{-}Hsiang Huang and
                  Caiming Xiong and
                  Chien{-}Sheng Wu},
  title        = {Agentic Uncertainty Quantification},
  journal      = {CoRR},
  volume       = {abs/2601.15703},
  year         = {2026},
  url          = {https://doi.org/10.48550/arXiv.2601.15703},
  doi          = {10.48550/ARXIV.2601.15703},
  eprinttype   = {arXiv},
  eprint       = {2601.15703},
  bibsource    = {dblp computer science bibliography, https://dblp.org}
}

@article{safeground2026,
  author       = {Qingni Wang and
                  Yue Fan and
                  Xin Eric Wang},
  title        = {SafeGround: Know When to Trust {GUI} Grounding Models via Uncertainty
                  Calibration},
  journal      = {CoRR},
  volume       = {abs/2602.02419},
  year         = {2026},
  url          = {https://doi.org/10.48550/arXiv.2602.02419},
  doi          = {10.48550/ARXIV.2602.02419},
  eprinttype   = {arXiv},
  eprint       = {2602.02419},
  bibsource    = {dblp computer science bibliography, https://dblp.org}
}

@article{hyperclick2025,
  author       = {Shaojie Zhang and
                  Pei Fu and
                  Ruoceng Zhang and
                  Jiahui Yang and
                  Anan Du and
                  Xiuwen Xi and
                  Shaokang Wang and
                  Ying Huang and
                  Bin Qin and
                  Zhenbo Luo and
                  Jian Luan},
  title        = {HyperClick: Advancing Reliable {GUI} Grounding via Uncertainty Calibration},
  journal      = {CoRR},
  volume       = {abs/2510.27266},
  year         = {2025},
  url          = {https://doi.org/10.48550/arXiv.2510.27266},
  doi          = {10.48550/ARXIV.2510.27266},
  eprinttype   = {arXiv},
  eprint       = {2510.27266},
  bibsource    = {dblp computer science bibliography, https://dblp.org}
}

@inproceedings{seeclick2024,
  author       = {Kanzhi Cheng and
                  Qiushi Sun and
                  Yougang Chu and
                  Fangzhi Xu and
                  Yantao Li and
                  Jianbing Zhang and
                  Zhiyong Wu},
  editor       = {Lun{-}Wei Ku and
                  Andre Martins and
                  Vivek Srikumar},
  title        = {SeeClick: Harnessing {GUI} Grounding for Advanced Visual {GUI} Agents},
  booktitle    = {Proceedings of the 62nd Annual Meeting of the Association for Computational
                  Linguistics (Volume 1: Long Papers), {ACL} 2024, Bangkok, Thailand,
                  August 11-16, 2024},
  pages        = {9313--9332},
  publisher    = {Association for Computational Linguistics},
  year         = {2024},
  url          = {https://doi.org/10.18653/v1/2024.acl-long.505},
  doi          = {10.18653/V1/2024.ACL-LONG.505},
  bibsource    = {dblp computer science bibliography, https://dblp.org}
}

@inproceedings{tao2025localization,
  author       = {Xingjian Tao and
                  Yiwei Wang and
                  Yujun Cai and
                  Zhicheng Yang and
                  Jing Tang},
  editor       = {Christos Christodoulopoulos and
                  Tanmoy Chakraborty and
                  Carolyn Rose and
                  Violet Peng},
  title        = {Understanding {GUI} Agent Localization Biases through Logit Sharpness},
  booktitle    = {Findings of the Association for Computational Linguistics: {EMNLP}
                  2025, Suzhou, China, November 4-9, 2025},
  pages        = {23361--23374},
  publisher    = {Association for Computational Linguistics},
  year         = {2025},
  url          = {https://aclanthology.org/2025.findings-emnlp.1268/},
  bibsource    = {dblp computer science bibliography, https://dblp.org}
}

@article{wen2025abstention,
  author       = {Bingbing Wen and
                  Jihan Yao and
                  Shangbin Feng and
                  Chenjun Xu and
                  Yulia Tsvetkov and
                  Bill Howe and
                  Lucy Lu Wang},
  title        = {Know Your Limits: {A} Survey of Abstention in Large Language Models},
  journal      = {Trans. Assoc. Comput. Linguistics},
  volume       = {13},
  pages        = {529--556},
  year         = {2025},
  url          = {https://doi.org/10.1162/tacl\_a\_00754},
  doi          = {10.1162/TACL\_A\_00754},
  bibsource    = {dblp computer science bibliography, https://dblp.org}
}

@article{snell2025tts,
  author       = {Charlie Snell and
                  Jaehoon Lee and
                  Kelvin Xu and
                  Aviral Kumar},
  title        = {Scaling {LLM} Test-Time Compute Optimally can be More Effective than
                  Scaling Model Parameters},
  journal      = {CoRR},
  volume       = {abs/2408.03314},
  year         = {2024},
  url          = {https://doi.org/10.48550/arXiv.2408.03314},
  doi          = {10.48550/ARXIV.2408.03314},
  eprinttype   = {arXiv},
  eprint       = {2408.03314},
  bibsource    = {dblp computer science bibliography, https://dblp.org}
}

@inproceedings{madaan2023selfrefine,
  author       = {Aman Madaan and
                  Niket Tandon and
                  Prakhar Gupta and
                  Skyler Hallinan and
                  Luyu Gao and
                  Sarah Wiegreffe and
                  Uri Alon and
                  Nouha Dziri and
                  Shrimai Prabhumoye and
                  Yiming Yang and
                  Shashank Gupta and
                  Bodhisattwa Prasad Majumder and
                  Katherine Hermann and
                  Sean Welleck and
                  Amir Yazdanbakhsh and
                  Peter Clark},
  editor       = {Alice Oh and
                  Tristan Naumann and
                  Amir Globerson and
                  Kate Saenko and
                  Moritz Hardt and
                  Sergey Levine},
  title        = {Self-Refine: Iterative Refinement with Self-Feedback},
  booktitle    = {Advances in Neural Information Processing Systems 36: Annual Conference
                  on Neural Information Processing Systems 2023, NeurIPS 2023, New Orleans,
                  LA, USA, December 10 - 16, 2023},
  year         = {2023},
  url          = {http://papers.nips.cc/paper\_files/paper/2023/hash/91edff07232fb1b55a505a9e9f6c0ff3-Abstract-Conference.html},
  bibsource    = {dblp computer science bibliography, https://dblp.org}
}

@article{farquhar2024semantic,
  author       = {Sebastian Farquhar and
                  Jannik Kossen and
                  Lorenz Kuhn and
                  Yarin Gal},
  title        = {Detecting hallucinations in large language models using semantic entropy},
  journal      = {Nat.},
  volume       = {630},
  number       = {8017},
  pages        = {625--630},
  year         = {2024},
  url          = {https://doi.org/10.1038/s41586-024-07421-0},
  doi          = {10.1038/S41586-024-07421-0},
  bibsource    = {dblp computer science bibliography, https://dblp.org}
}

@article{trustworthygui2025,
  author       = {Yucheng Shi and
                  Wenhao Yu and
                  Wenlin Yao and
                  Wenhu Chen and
                  Ninghao Liu},
  title        = {Towards Trustworthy {GUI} Agents: {A} Survey},
  journal      = {CoRR},
  volume       = {abs/2503.23434},
  year         = {2025},
  url          = {https://doi.org/10.48550/arXiv.2503.23434},
  doi          = {10.48550/ARXIV.2503.23434},
  eprinttype   = {arXiv},
  eprint       = {2503.23434},
  bibsource    = {dblp computer science bibliography, https://dblp.org}
}

@inproceedings{aguvis2024,
  author       = {Yiheng Xu and
                  Zekun Wang and
                  Junli Wang and
                  Dunjie Lu and
                  Tianbao Xie and
                  Amrita Saha and
                  Doyen Sahoo and
                  Tao Yu and
                  Caiming Xiong},
  editor       = {Aarti Singh and
                  Maryam Fazel and
                  Daniel Hsu and
                  Simon Lacoste{-}Julien and
                  Felix Berkenkamp and
                  Tegan Maharaj and
                  Kiri Wagstaff and
                  Jerry Zhu},
  title        = {Aguvis: Unified Pure Vision Agents for Autonomous {GUI} Interaction},
  booktitle    = {Forty-second International Conference on Machine Learning, {ICML}
                  2025, Vancouver, BC, Canada, July 13-19, 2025},
  series       = {Proceedings of Machine Learning Research},
  publisher    = {{PMLR} / OpenReview.net},
  year         = {2025},
  url          = {https://proceedings.mlr.press/v267/xu25ae.html},
  bibsource    = {dblp computer science bibliography, https://dblp.org}
}

@article{osatlas2024,
  author       = {Zhiyong Wu and
                  Zhenyu Wu and
                  Fangzhi Xu and
                  Yian Wang and
                  Qiushi Sun and
                  Chengyou Jia and
                  Kanzhi Cheng and
                  Zichen Ding and
                  Liheng Chen and
                  Paul Pu Liang and
                  Yu Qiao},
  title        = {{OS-ATLAS:} {A} Foundation Action Model for Generalist {GUI} Agents},
  journal      = {CoRR},
  volume       = {abs/2410.23218},
  year         = {2024},
  url          = {https://doi.org/10.48550/arXiv.2410.23218},
  doi          = {10.48550/ARXIV.2410.23218},
  eprinttype   = {arXiv},
  eprint       = {2410.23218},
  bibsource    = {dblp computer science bibliography, https://dblp.org}
}

@inproceedings{rawles2024androidworld,
  author       = {Christopher Rawles and
                  Sarah Clinckemaillie and
                  Yifan Chang and
                  Jonathan Waltz and
                  Gabrielle Lau and
                  Marybeth Fair and
                  Alice Li and
                  William E. Bishop and
                  Wei Li and
                  Folawiyo Campbell{-}Ajala and
                  Daniel Kenji Toyama and
                  Robert James Berry and
                  Divya Tyamagundlu and
                  Timothy P. Lillicrap and
                  Oriana Riva},
  title        = {AndroidWorld: {A} Dynamic Benchmarking Environment for Autonomous
                  Agents},
  booktitle    = {The Thirteenth International Conference on Learning Representations,
                  {ICLR} 2025, Singapore, April 24-28, 2025},
  publisher    = {OpenReview.net},
  year         = {2025},
  url          = {https://openreview.net/forum?id=il5yUQsrjC},
  bibsource    = {dblp computer science bibliography, https://dblp.org}
}

@inproceedings{lin2024showui,
  author       = {Kevin Qinghong Lin and
                  Linjie Li and
                  Difei Gao and
                  Zhengyuan Yang and
                  Shiwei Wu and
                  Zechen Bai and
                  Stan Weixian Lei and
                  Lijuan Wang and
                  Mike Zheng Shou},
  title        = {ShowUI: One Vision-Language-Action Model for {GUI} Visual Agent},
  booktitle    = {{IEEE/CVF} Conference on Computer Vision and Pattern Recognition,
                  {CVPR} 2025, Nashville, TN, USA, June 11-15, 2025},
  pages        = {19498--19508},
  publisher    = {Computer Vision Foundation / {IEEE}},
  year         = {2025},
  url          = {https://openaccess.thecvf.com/content/CVPR2025/html/Lin\_ShowUI\_One\_Vision-Language-Action\_Model\_for\_GUI\_Visual\_Agent\_CVPR\_2025\_paper.html},
  doi          = {10.1109/CVPR52734.2025.01816},
  bibsource    = {dblp computer science bibliography, https://dblp.org}
}

@inproceedings{zheng2024seeact,
  author       = {Boyuan Zheng and
                  Boyu Gou and
                  Jihyung Kil and
                  Huan Sun and
                  Yu Su},
  editor       = {Ruslan Salakhutdinov and
                  Zico Kolter and
                  Katherine A. Heller and
                  Adrian Weller and
                  Nuria Oliver and
                  Jonathan Scarlett and
                  Felix Berkenkamp},
  title        = {GPT-4V(ision) is a Generalist Web Agent, if Grounded},
  booktitle    = {Forty-first International Conference on Machine Learning, {ICML} 2024,
                  Vienna, Austria, July 21-27, 2024},
  series       = {Proceedings of Machine Learning Research},
  pages        = {61349--61385},
  publisher    = {{PMLR} / OpenReview.net},
  year         = {2024},
  url          = {https://proceedings.mlr.press/v235/zheng24e.html},
  bibsource    = {dblp computer science bibliography, https://dblp.org}
}

@inproceedings{geifman2017selective,
  author       = {Yonatan Geifman and
                  Ran El{-}Yaniv},
  editor       = {Isabelle Guyon and
                  Ulrike von Luxburg and
                  Samy Bengio and
                  Hanna M. Wallach and
                  Rob Fergus and
                  S. V. N. Vishwanathan and
                  Roman Garnett},
  title        = {Selective Classification for Deep Neural Networks},
  booktitle    = {Advances in Neural Information Processing Systems 30: Annual Conference
                  on Neural Information Processing Systems 2017, December 4-9, 2017,
                  Long Beach, CA, {USA}},
  pages        = {4878--4887},
  year         = {2017},
  url          = {https://proceedings.neurips.cc/paper/2017/hash/4a8423d5e91fda00bb7e46540e2b0cf1-Abstract.html},
  bibsource    = {dblp computer science bibliography, https://dblp.org}
}

@inproceedings{malinin2021uncertainty,
  author       = {Andrey Malinin and
                  Mark J. F. Gales},
  title        = {Uncertainty Estimation in Autoregressive Structured Prediction},
  booktitle    = {9th International Conference on Learning Representations, {ICLR} 2021,
                  Virtual Event, Austria, May 3-7, 2021},
  publisher    = {OpenReview.net},
  year         = {2021},
  url          = {https://openreview.net/forum?id=jN5y-zb5Q7m},
  bibsource    = {dblp computer science bibliography, https://dblp.org}
}

@inproceedings{uground2025,
  author       = {Boyu Gou and
                  Ruohan Wang and
                  Boyuan Zheng and
                  Yanan Xie and
                  Cheng Chang and
                  Yiheng Shu and
                  Huan Sun and
                  Yu Su},
  title        = {Navigating the Digital World as Humans Do: Universal Visual Grounding
                  for {GUI} Agents},
  booktitle    = {The Thirteenth International Conference on Learning Representations,
                  {ICLR} 2025, Singapore, April 24-28, 2025},
  publisher    = {OpenReview.net},
  year         = {2025},
  url          = {https://openreview.net/forum?id=kxnoqaisCT},
  bibsource    = {dblp computer science bibliography, https://dblp.org}
}

\appendix
\section{PACE Algorithm and Applicability}
  \label{app:algorithm}

  \begin{figure}[H]
  \centering
  \fbox{\begin{minipage}{0.93\columnwidth}
  \small
  \textbf{Algorithm 1: Place-Aware Coordinate Entropy (PACE)}\\[2pt]
  \textbf{Input:} screenshot $I$, instruction $q$, model $M$, exponent $\alpha$\\
  \textbf{Output:} confidence score $c \in \mathbb{R}$, where larger values indicate higher confidence\\[2pt]

  \quad 1: $y_{1:T}, z_{1:T} \gets M.\mathrm{generate}(I,q)$
  \hfill (greedy decoding with logits)\\

  \quad 2: $K_x,K_y \gets \mathrm{coord\_digit\_indices}(y_{1:T})$\\

  \quad 3: \textbf{if} no valid fixed-scale digit-place map exists \textbf{then}\\

  \quad 4: \quad \textbf{return} negative uniform Coordinate-Entropy\\

  \quad 5: \textbf{for} each axis $a\in\{x,y\}$ and digit position
  $j\in\{1,\ldots,L_a\}$ \textbf{do}\\

  \quad 6: \quad $p_{a,j}\gets\mathrm{softmax}(z_{a,j})$\\

  \quad 7: \quad $H_{a,j}\gets-\sum_v p_{a,j}(v)\log p_{a,j}(v)$\\

  \quad 8: \quad $v_{a,j}\gets10^{L_a-j}$\\

  \quad 9: \textbf{return}
  $c_\alpha
  =-\displaystyle\frac{\sum_{a,j}v_{a,j}^{\alpha}H_{a,j}}
  {\sum_{a,j}v_{a,j}^{\alpha}}$
  \end{minipage}}
  \caption{\textbf{PACE computation.} PACE uses coordinate-token logits from the same greedy decode that produces the click and therefore requires no additional
  forward pass.}
  \label{alg:pwce}
  \end{figure}

  PACE requires each coordinate digit to be emitted as a separate token on a stable coordinate scale. Our primary evaluation uses outputs with exactly three digit
  tokens on each axis. When this condition is not satisfied, as with variable-scale pixels, fractional coordinates, or merged digit tokens, the implementation falls
  back to uniform Coordinate-Entropy.

\section{Experimental Details}
  \label{app:details}

  \paragraph{Datasets.}
  Table~\ref{tab:appendix-datasets} summarizes the two evaluation benchmarks. PACE requires no training, and we use the fixed $\alpha{=}1$ setting throughout.

  \begin{table}[H]
  \centering
  \small
  \setlength{\tabcolsep}{3pt}
  \renewcommand{\arraystretch}{1.10}
  \resizebox{\columnwidth}{!}{%
  \begin{tabular}{@{}lp{0.32\columnwidth}rll@{}}
  \toprule
  \textbf{Dataset}
  & \textbf{Breakdown}
  & \textbf{Total}
  & \textbf{Citation}
  & \textbf{License} \\
  \midrule
  ScreenSpot-Pro
  & $23$ application categories across $3$ operating systems, covering professional desktop interfaces
  & $1{,}581$
  & \citet{screenspotpro2025}
  & MIT \\
  \midrule
  ScreenSpot-v$2$
  & $8$ platforms covering consumer mobile, web, and desktop interfaces
  & $1{,}272$
  & \citet{osatlas2024}
  & Apache $2.0$ \\
  \bottomrule
  \end{tabular}}
  \caption{\textbf{Evaluation benchmarks.} Both datasets contain an instruction, a screenshot, and a target bounding box. A predicted click is correct if it falls
  inside the target box. The primary click is greedily decoded; additional stochastic generations are used only by sampling-based confidence baselines.}
  \label{tab:appendix-datasets}
  \end{table}
  \paragraph{Inference parameters.}
  Table~\ref{tab:appendix-inference} summarizes the decoding configuration for each confidence-signal family.

  \begin{table}[H]
  \centering
  \small
  \setlength{\tabcolsep}{4pt}
  \renewcommand{\arraystretch}{1.10}
  \resizebox{\columnwidth}{!}{%
  \begin{tabular}{@{}lll@{}}
  \toprule
  \textbf{Signal family}
  & \textbf{Decoding}
  & \textbf{Confidence computation} \\
  \midrule
  PACE, CE, and full-sequence
  & Greedy, $T{=}0$
  & Token logits from one decode \\
  SafeGround
  & $K{=}10$, $T{=}1.0$
  & Uncertainty from sampled clicks \\
  Verbalized
  & Greedy meta-response
  & Confidence extracted from the meta-prompt response \\
  \bottomrule
  \end{tabular}}
  \caption{\textbf{Inference configuration by confidence-signal family.} Sampling and verbalized methods estimate confidence for the same greedily decoded primary
  click used by the single-pass methods. Models are evaluated in bfloat16 on NVIDIA H$100$ or RTX $3090$ GPUs.}
  \label{tab:appendix-inference}
  \end{table}

  \paragraph{Coordinate-token identification.}
  For structured tool calls, we locate the coordinate field and align its decoded characters with the corresponding generation steps. We retain only tokens that emit
  a single decimal digit and parse the $x$ and $y$ axes separately. A prediction enters the fixed-three-digit analysis only when both axes contain exactly three such
  digit tokens. Tokens that merge multiple digits do not provide a valid place map and therefore fall back to Coordinate-Entropy.

\section{Structural Properties: Coordinate vs.\ Full-Sequence Tokens}
  \label{app:structural_properties}

  Coordinate-Entropy outperforms full-sequence logit summaries because the two approaches aggregate uncertainty from different parts of the response.

  \paragraph{Action relevance.}
  Full responses contain reasoning, formatting, and tool-call tokens whose variation does not necessarily change the predicted click. Coordinate tokens are more
  directly tied to the executed action because changing a coordinate digit changes the predicted location. Although multiple coordinates within a target box may be
  correct, uncertainty at coordinate positions is therefore more closely related to click correctness than uncertainty in surrounding response tokens.

  \paragraph{Dilution under full-sequence aggregation.}
  Full-sequence summaries average coordinate uncertainty together with uncertainty from the rest of the response, which can dilute the signal associated with the
  predicted location. In contrast, Coordinate-Entropy evaluates only the tokens that determine the click. The strong performance of coordinate-level entropy and the
  coordinate-level top-two logit gap in Table~\ref{tab:headline} is consistent with this explanation.

\section{Sensitivity to the Place-Weight Exponent}
  \label{app:alpha_finegrid}

\begin{figure*}[!t]
  \centering
  \includegraphics[width=\textwidth]{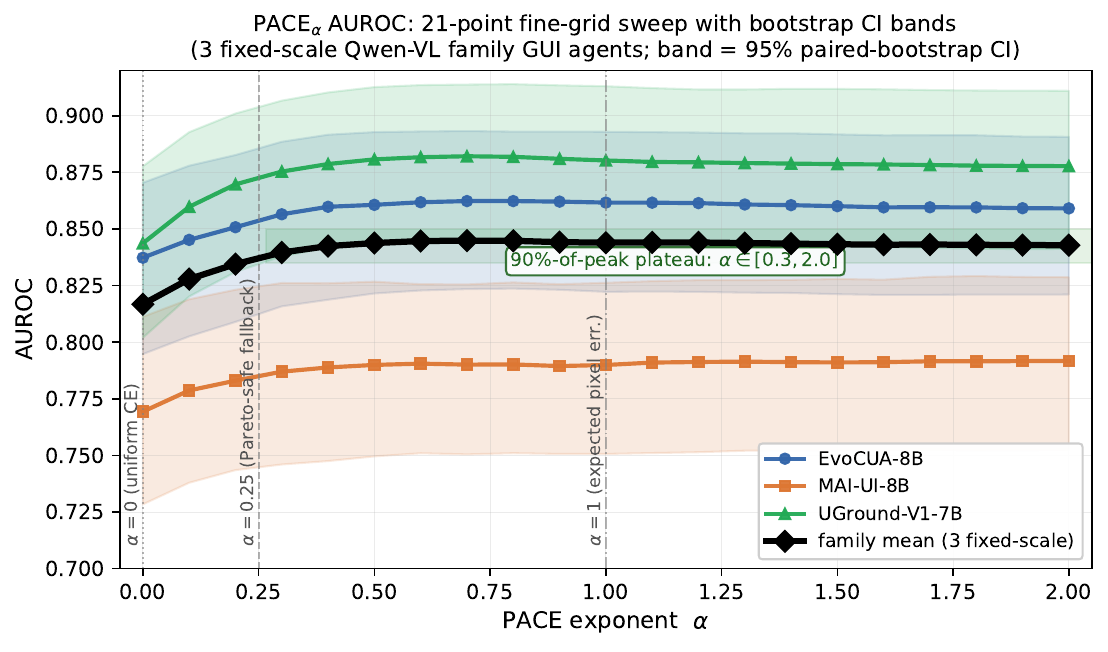}
  \caption{\textbf{Sensitivity to the place-weight exponent.}
  AUROC across three fixed-scale agents as $\alpha$ varies from $0$ to $2$, with $95\%$ paired-bootstrap confidence intervals.}
  \label{fig:alpha_finegrid}
  \end{figure*}

  We sweep $\alpha$ from $0$ to $2$ in increments of $0.1$ for the three primary fixed-scale agents: EvoCUA-$8$B, MAI-UI-$8$B, and UGround-V$1$-$7$B. As shown in
  Figure~\ref{fig:alpha_finegrid}, PACE$_{\alpha{=}1}$ lies within the high-AUROC region for all three agents. The mean curve remains within $0.01$ of its maximum
  throughout $\alpha \in [0.3,2.0]$, indicating that performance is not sensitive to the exact exponent around the default setting.

\section{Statistical Significance of Place Weighting}
  \label{app:significance}

  We evaluate the AUROC difference between PACE$_{\alpha{=}1}$ and
  Coordinate-Entropy using $2{,}000$ paired-bootstrap resamples. Predictions
  from the two methods are paired on the same examples within each agent and
  benchmark. As shown in Table~\ref{tab:pace_ce_significance}, all six point
  estimates favor PACE, and four have confidence intervals that exclude zero.

  \begin{table}[H]
  \centering
  \small
  \setlength{\tabcolsep}{4pt}
  \resizebox{\columnwidth}{!}{%
  \begin{tabular}{lcc}
  \toprule
  \textbf{Agent}
  & \textbf{ScreenSpot-Pro}
  & \textbf{ScreenSpot-v$2$} \\
  \midrule
  EvoCUA-$8$B
  & $+0.010\;[-0.007,\,+0.026]$
  & $\mathbf{+0.065\;[+0.018,\,+0.111]}$ \\
  MAI-UI-$8$B
  & $\mathbf{+0.031\;[+0.009,\,+0.052]}$
  & $+0.015\;[-0.041,\,+0.070]$ \\
  UGround-V$1$-$7$B
  & $\mathbf{+0.025\;[+0.013,\,+0.037]}$
  & $\mathbf{+0.054\;[+0.027,\,+0.080]}$ \\
  \bottomrule
  \end{tabular}}
  \caption{\textbf{Paired AUROC differences between PACE and
  Coordinate-Entropy.} Each entry reports
  $\Delta\mathrm{AUROC}=\mathrm{PACE}-\mathrm{CE}$ with a $95\%$
  paired-bootstrap confidence interval. Bold indicates that the interval
  excludes zero.}
  \label{tab:pace_ce_significance}
  \end{table}

\section{Fine-Grained Localization}
  \label{app:finegrained}

  \paragraph{Performance across localization-error distances.}
  We divide incorrect clicks into near, intermediate, and far localization
  errors, and compare each error group against all correct clicks. This
  analysis complements the same-hundreds result in the main text by showing
  where place weighting helps and where uniform Coordinate-Entropy remains
  competitive.

  \begin{table}[H]
  \centering
  \small
  \setlength{\tabcolsep}{4pt}
  \resizebox{\columnwidth}{!}{%
  \begin{tabular}{lccc}
  \toprule
  & \multicolumn{3}{c}{\textbf{PACE / Coordinate-Entropy AUROC}} \\
  \cmidrule(lr){2-4}
  \textbf{Agent}
  & \textbf{Near ($<50$)}
  & \textbf{Intermediate ($50$--$150$)}
  & \textbf{Far ($\geq150$)} \\
  \midrule
  EvoCUA-$8$B
  & $0.789 / \mathbf{0.832}$
  & $\mathbf{0.872} / 0.854$
  & $\mathbf{0.920} / 0.858$ \\
  MAI-UI-$8$B
  & $\mathbf{0.706} / 0.694$
  & $\mathbf{0.886} / 0.865$
  & $\mathbf{0.876} / 0.811$ \\
  UGround-V$1$-$7$B
  & $0.779 / \mathbf{0.802}$
  & $0.892 / 0.892$
  & $\mathbf{0.950} / 0.900$ \\
  \bottomrule
  \end{tabular}}
  \caption{\textbf{Discrimination across localization-error distances on
  ScreenSpot-Pro.} Incorrect clicks are grouped by their pixel-distance
  error, and each group is compared against all correct clicks. Bold marks
  the higher AUROC within each pair.}
  \label{tab:error_distance}
  \end{table}

  PACE provides its clearest improvement for larger localization errors,
  outperforming Coordinate-Entropy for all three agents in the far-error
  group. The comparison is mixed for near errors, where Coordinate-Entropy
  performs better for two of the three agents. This pattern is consistent
  with the place-value mechanism: coarse-position uncertainty is most
  useful for identifying large displacements, while finer uncertainty
  remains relevant for clicks close to the target.

  \section{AUPRC Results}
  \label{app:auprc}

  We additionally report the area under the precision--recall curve
  (AUPRC), which summarizes the precision--recall tradeoff as the
  error-detection threshold varies. We treat an incorrect click as the
  positive class and use negative confidence as the error score.
  PACE$_{\alpha{=}1}$ has higher AUPRC than Coordinate-Entropy in all six
  primary agent--benchmark comparisons, consistent with the AUROC results.

  \begin{table}[H]
  \centering
  \small
  \setlength{\tabcolsep}{4pt}
  \resizebox{\columnwidth}{!}{%
  \begin{tabular}{lcc|cc}
  \toprule
  & \multicolumn{2}{c|}{\textbf{ScreenSpot-Pro}}
  & \multicolumn{2}{c}{\textbf{ScreenSpot-v$2$}} \\
  \textbf{Agent}
  & \textbf{CE}
  & \textbf{PACE}
  & \textbf{CE}
  & \textbf{PACE} \\
  \midrule
  EvoCUA-$8$B
  & $0.835$ & $\mathbf{0.841}$
  & $0.214$ & $\mathbf{0.309}$ \\
  MAI-UI-$8$B
  & $0.722$ & $\mathbf{0.733}$
  & $0.078$ & $\mathbf{0.107}$ \\
  UGround-V$1$-$7$B
  & $0.952$ & $\mathbf{0.963}$
  & $0.416$ & $\mathbf{0.502}$ \\
  \bottomrule
  \end{tabular}}
  \caption{\textbf{AUPRC with incorrect clicks as the positive class.}
  PACE and Coordinate-Entropy are evaluated on the same coordinate-output
  subset within each cell. Because AUPRC depends on error prevalence,
  values should be compared within, rather than across, agent--benchmark
  cells.}
  \label{tab:auprc}
  \end{table}
  
\section{Emission-Format Boundaries}
  \label{app:format_boundary}

  \paragraph{Four-digit fixed-screen coordinates.}
  We evaluate whether place weighting extends beyond three-digit
  coordinates using UI-TARS-1.5 on AndroidControl.
  This setting contains $4{,}916$ click steps on a fixed
  $924{\times}2100$ screen. The thousands place is nearly constant because
  it is constrained by the screen geometry and therefore carries little
  information about click correctness. Table~\ref{tab:four_digit} compares
  uniform Coordinate-Entropy, literal four-place weighting, and a
  format-aware variant that omits the geometry-pinned thousands place.

  \begin{table}[H]
  \centering
  \small
  \setlength{\tabcolsep}{4pt}
  \resizebox{\columnwidth}{!}{%
  \begin{tabular}{lcc}
  \toprule
  \textbf{Weighting scheme}
  & \textbf{Relative weights}
  & \textbf{$\Delta$AUROC [95\% CI]} \\
  \midrule
  Coordinate-Entropy
  & $1{:}1{:}1{:}1$
  & $0.000$ \\
  Literal four-place PACE
  & $1000{:}100{:}10{:}1$
  & $+0.020\;[+0.001,\,+0.039]$ \\
  Format-aware PACE
  & $0{:}100{:}10{:}1$
  & $\mathbf{+0.061\;[+0.048,\,+0.076]}$ \\
  \bottomrule
  \end{tabular}}
  \caption{\textbf{Four-digit coordinate weighting for UI-TARS-1.5 on
  AndroidControl.} Results use $4{,}916$ click steps collected on a fixed
  $924{\times}2100$ screen. AUROC differences are measured relative to
  uniform Coordinate-Entropy. The format-aware variant omits the geometry-constrained thousands place.}
  \label{tab:four_digit}
  \end{table}

  Literal place weighting already improves over uniform
  Coordinate-Entropy in this fixed-screen setting. The larger gain from
  the format-aware variant shows that a leading digit should not receive
  high weight when its variation is determined primarily by screen
  geometry rather than click location.

  \paragraph{Other emission formats.}
  Place weighting requires a stable correspondence between an emitted
  digit position and its geometric displacement on the screen.
  Table~\ref{tab:format_boundary} summarizes the evaluated formats.
  Fixed-scale integer coordinates satisfy this requirement directly.
  When coordinate length, resolution, or tokenization removes this
  correspondence, uniform Coordinate-Entropy is the safer default.

  \begin{table*}[!t]
  \centering
  \small
  \setlength{\tabcolsep}{4pt}
  \begin{tabular}{
  p{0.20\textwidth}
  p{0.17\textwidth}
  p{0.29\textwidth}
  p{0.22\textwidth}}
  \toprule
  \textbf{Emission format}
  & \textbf{Example}
  & \textbf{Observed behavior}
  & \textbf{Recommended signal} \\
  \midrule
  Fixed-scale, three-digit integer
  & UI-TARS-$72$B-DPO
  & PACE improves over Coordinate-Entropy on both benchmarks.
  & PACE with standard place-value weights. \\
  \midrule
  Fixed-screen, four-digit integer
  & UI-TARS-1.5 on AndroidControl
  & Literal four-place weighting gives $+0.020$ AUROC; omitting the
  geometry-pinned leading digit gives $+0.061$.
  & PACE with a format-aligned digit window. \\
  \midrule Variable-resolution pixel coordinates
  & Qwen2.5-VL-$32$B
  & Fixed place weighting gives $-0.012/-0.025$ AUROC relative to
  Coordinate-Entropy.
  & Coordinate-Entropy unless a stable geometric mapping is available. \\
  \midrule
  Short coordinate emission
  & Magma-$8$B
  & PACE approximately matches Coordinate-Entropy
  ($-0.005/-0.010$).
  & Coordinate-Entropy. \\
  \midrule
  Fractional coordinates in $[0,1]$
  & Aguvis-$7$B
  & PACE remains close to Coordinate-Entropy
  ($-0.006/+0.018$).
  & Coordinate-Entropy unless the informative precision is known. \\
  \midrule
  Merged coordinate tokens
  & CogAgent-$9$B
  & Individual digit positions cannot be identified because multiple
  digits form one token.
  & Coordinate-token entropy without digit-level weighting. \\
  \bottomrule
  \end{tabular}
  \caption{\textbf{Applicability across coordinate-emission formats.}
  Slash-separated differences report ScreenSpot-Pro and ScreenSpot-v$2$,
  respectively. PACE is used when digit positions have a stable geometric
  meaning; otherwise the method falls back to uniform Coordinate-Entropy.}
  \label{tab:format_boundary}
  \end{table*}

  These results define the scope of the default formulation. PACE applies
  directly to separately tokenized, fixed-scale integer coordinates and
  can be extended to a fixed four-digit format by aligning the weighted
  window with the varying digit positions. For variable-scale,
  fractional, or merged-token formats, we use Coordinate-Entropy rather
  than imposing an unsupported place mapping.

\end{document}